\documentclass{ws-ijufks}

\usepackage[utf8]{inputenc}
\usepackage{elocalloc}
\usepackage{tikz}
\usetikzlibrary{calc,positioning,decorations.pathreplacing}
\usepackage{enumitem}
\usepackage{forest}

\usepackage{graphicx}
\usepackage{url}
\usepackage{alltt}

\newcommand{\tr}[1]{\hbox{\bfseries\sffamily#1}}

\newcommand{\trZ}[1]{\ensuremath{{}^0\mkern-0.8mu\hbox{\sf#1}}}

\newcommand{\trZcon}[1]{\ensuremath{{}^0\mkern-0.8mu#1}}

\renewcommand{\t}{\ensuremath{\tau}} 

\newcommand{\om}{\ensuremath{\omega}}
\newcommand{\tw}[1]{\ensuremath{{#1}_{\t\om}}}

\newcommand{\tstar}[1][n]{\ensuremath{*_{#1}}}

\newcommand{\term}[2]{%
    \begin{tabular}{c}\hbox{\textcolor{blue}{{\scriptsize\bf#1}}}\\
    \hbox{\textcolor{magenta}{{\tiny\bf#2}}}\end{tabular}}

\begin{document}

\markboth{M. Duží and A. Horák}{Hyperintensional Reasoning based on NL Knowledge Base}

%
\catchline{}{}{}{}{}
%

\title{HYPERINTENSIONAL REASONING BASED ON NATURAL LANGUAGE KNOWLEDGE BASE}

\author{MARIE DUŽÍ}

\address{VSB-Technical University Ostrava, Department of Computer Science FEI \\
       17. listopadu 15, 708 33 Ostrava, Czech Republic \\
       marie.duzi@vsb.cz}           

\author{ALEŠ HORÁK}

\address{Natural Language Processing Centre\\
       Faculty of Informatics, Masaryk University\\
       Botanicka 68a, 602 00, Brno, Czech Republic\\
       hales@fi.muni.cz}

\maketitle

\begin{history}
\received{(received date)}
\revised{(revised date)}
\end{history}

\begin{abstract}
The success of automated reasoning techniques over large
natural-language texts heavily relies on a fine-grained analysis of
natural language assumptions. While there is a common agreement that
the analysis should be hyperintensional, most of the automatic
reasoning systems are still based on an intensional logic, at the
best. In this paper, we introduce the system of reasoning based on
a fine-grained, hyperintensional analysis. To this end we apply
Tichy's Transparent Intensional Logic (TIL) with its procedural
semantics. TIL is a higher-order, hyperintensional logic of partial
functions,  in particular apt for a fine-grained natural-language
analysis. Within TIL we recognise three kinds of context, namely
extensional, intensional and hyperintensional, in which a particular
natural-language term, or rather its meaning, can occur. Having
defined the three kinds of context and implemented an algorithm of
context recognition, we are in a position to develop and implement an
extensional logic of hyperintensions with the inference machine that
should neither over-infer nor under-infer.
\end{abstract}

\keywords{transparent intensional logic; hyperintensional logic;
natural language analysis; context recognition; knowledge based
system} 

\section{Introduction}
\label{sec:intro}

The family of automatic theorem provers, known today as HOL, is
getting more and more interest in logic, mathematics and computer
science (see, for instance, \cite{HOL2,HOL1,HOL3}).
These tools are broadly used in automatic theorem checking and applied
as interactive proof assistants. As `HOL' is an acronym for
higher-order logic, the underlying logic is usually a version of
a simply typed $\lambda$-calculus. This makes it possible to operate
both in extensional and intensional contexts, where a \emph{value} of
the denoted function or the \emph{function} itself, respectively, is
an object of predication.

Yet there is another application that is gaining interest, namely
reasoning over natural language statements. There are large amounts of
text data that we need to analyse and
formalize.\cite{reyes2017towards} Not only that, we also want to have
question-answer systems which would infer implicit computable
knowledge from these large explicit knowledge bases. To this end not
only intensional but rather \emph{hyperintensional} logic (see
\cite{SynthIntro}) is needed, because we need to formally analyse
natural language in a fine-grained way so that the underlying
inference machine is neither over-inferring (that yields
inconsistencies) nor under-inferring (that causes lack of knowledge).
We need to properly analyse agents' attitudes like knowing, believing,
seeking, solving, designing, etc., because attitudinal sentences are
part and parcel of our everyday vernacular. 

Since the substitution of logically equivalent clause for an attitude
complement can fail, as already Carnap in~\cite{Carnap} knew, we
need a fine-grained, hyperintensional analysis here.\footnote{Carnap
in \cite{Carnap} says that the complements of belief
attitudes are neither extensional not intensional. The term
‘hyperintensional’ was introduced by Creswell \cite{Cresswell} in
order to distinguish hyperintensional contexts from coarse-grained
extensional or intensional ones. Before possible-world semantics
occupied the term ‘intensional’ for extensionally-individuated
functions with the domain in possible world, the term ‘intensional’
was used in the same sense as the current term ‘hyperintensional’, in
particular in mathematics.} Thus the main reason for introducing
hyperintensional contexts was originally to block various invalid
interences, and hyperintensional contexts were defined in a negative
way, namely as those contexts that do not validate the substitution
of equivalent terms denoting the same object (see, e.g.
\cite{HyperDJ}). For instance, if Tilman (explicitly) believes that
the Pope is wise he does not have to believe that the Bishop of Rome
is wise, though both ‘the Pope’ and ‘the Bishop of Rome’ denote one
and the same papal office. Or, when Tilman computes $2+5$ he does not
compute $\sqrt{49}$ though both ‘2+5’ and ‘$\sqrt{49}$’ denote the
same number 7. He is trying to execute the procedure specified by the
term `2+5' rather than by `$\sqrt{49}$'. 

Yet, there is the other side of the coin, which is the positive topic
of which inferences should be validated in hyperintensional contexts.
For instance, if Tilman computes $Cotg(\pi)$ then there is a number
$x$ such that Tilman computes $Cotg(x)$. Our background theory is
Transparent Intensional Logic, or TIL for short (see
\cite{RefTIL2010,Tichy1988}), with input transformed directly from
natural language sentences by means of the Normal Translation
Algorithm \cite{RefHorak2002}. TIL definition of hyperintensionality
is \emph{positive} rather than negative. Any context in which the
meaning of an expression is \emph{displayed} rather than executed is
hyperintensional.  Moreover, our conception of meaning is
\emph{procedural}. Hyperintensions are abstract procedures rigorously
defined as TIL constructions which are assigned to expressions as
their context-invariant meanings. This entirely anti-contextual and
compositional semantics is, to the best of our knowledge, one of rare
theories that deals with all kinds of context, whether extensional,
intensional or hyperintensional, in a uniform way.\footnote{For the
disquotation theory of attitudinal sentences see, for instance,
\cite{Attitudes}, where general representation scheme for embedded
propositional content is presented. It is however just a scheme,
without a full semantic theory.} The same extensional logical laws are
valid invariably in all kinds of context. In particular, there is no
reason why Leibniz's law of substitution of identicals, and the rule
of existential generalisation were not valid. What differ according to
the context are not the rules themselves but the types of objects on
which these rules are validly applicable.  In an extensional context,
the \emph{value} of the function denoted by a given term is an object
of predication; hence, procedures producing the same value are
mutually substitutable. In an intensional context the denoted
\emph{function} itself is an object of predication; hence, procedures
producing the same function are mutually substitutable. Finally, in
a hyperintensional context the procedure that is the meanings itself
is an object of predication; thus only synonymous terms encoding the
same procedure are substitutable. Due to its stratified ontology of
entities organised in a ramified hierarchy of types, TIL is a logical
framework within which such an extensional logic of hyperintensions
has been introduced, see \cite{Duz1,Duz2}.

Reasoning over natural language is usually limited to the
\emph{textual entailment} task \cite{RefBentivogli2011}, which relies
on lexical knowledge transfer and stochastic relations rather than
deriving logical consequences of facts encoded in these texts. In this
paper, we are going to fill this gap. We introduce a system for
reasoning based on  TIL with natural language input. The task of
\emph{logical analysis} of natural language sentences is connected
with advanced formalisms for natural language syntactic analysis, such
as the Head-driven Phrase Structure Grammar \cite{RefCarroll2004} or
Combinatory Categorial Grammars
\cite{RefKwiatkowski2011,RefZettlemoyer2012}. Even though these
formalisms make it possible to transform an input sentence into
a logical representation, the expressiveness of the logical formalism
is usually within limits of the first-order predicate logic or
descriptive logic \cite{RefCeylan2017,MartinezSantiago2017}.  In this
paper we introduce an algorithm that combines syntactic analysis with
logical analysis based on the \emph{procedural}, i.e. hyperintensional
semantics of TIL. This algorithm exploits the TIL type lexicon of
10,500 verb-type assignments and about 30,000 logical schemata for
verbs that serve for assigning correct types to verb arguments thus
making it possible to discover a fine-grained meaning encoded in the
form of a TIL construction. Furthermore, we make use of VerbaLex
lexicon of Czech verb valencies containing deep verb frames. These
frames are then used to propose TIL types assigned to verbs and verb
logical schemata. In order to assign types to verb arguments, we
exploit the links to Princeton WordNet. Currently we have the corpus
of more than 6,000 TIL constructions that serve for computer-aided
analysis of language.

The rest of the paper is organised as follows. In
Section~\ref{sec:til} we introduce the basic principles of TIL, the
algorithm of context recognition, substitution method and
\mbox{$\beta$-conversion} `by value'. Last but not least, here we also
deal with sentences that come attached with a presupposition.
Section~\ref{sec:loganal} deals with the method of automatic
processing natural language texts. In Section~\ref{sec:reason} we
introduce reasoning in TIL based on a variant of the general
resolution method and demonstrate its principles by a simple example.
Finally, concluding remarks are presented in Section~\ref{sec:concl}.

\section{Fundamentals of TIL}
\label{sec:til}

The TIL syntax will be familiar to those who are familiar with the
syntax of $\lambda$-calculi with four important
exceptions.\footnote{For details see  
\cite{RefTIL2010}.}

First, TIL $\lambda$-terms denote abstract procedures rigorously defined as
constructions, rather than the set-theoretic functions produced by
these procedures. Thus, the construction Composition symbolised by $[F
A_1 ... A_m]$ is the very procedure of applying a function presented
by $F$ to an argument presented by $A_1, ..., A_m$, and the
construction Closure $[\lambda x_1 x_2 ... x_n C]$ is the very
procedure of constructing a function by \hbox{$\lambda$-abstraction}
in the ordinary manner of $\lambda$-calculi. Second, objects to be
operated on by complex constructions must be supplied by atomic
constructions. Atomic constructions are one-step procedures that do
not contain any other constituents but themselves. They are
\emph{variables} and \emph{Trivialization}. Variables construct
entities of the respective types dependently on valuation, they
\hbox{$v$-construct}. For each type (see Def. 2) there are countably
many variables assigned that range over this type ($v$-construct
entities of this type). Trivialisation $^0\!X$ of an entity $X$ (of
any type even a construction) constructs simply $X$. In order to
operate on an entity $X$, the entity must be grabbed first.
Trivialisation is such a one-step grabbing mechanism.\footnote{In our
system Trivialization is implemented as a pointer to $X$.}  Third,
since the product of a construction can be another construction,
constructions can be executed twice over. To this end we have Double
Execution of $X$, $^2\!X$, that $v$-constructs what is $v$-constructed
by the product of $X$. Finally, since we work with partial functions,
constructions can be $v$-improper in the sense of failing to
$v$-construct an object for a valuation $v$.

\subsection{Constructions and types}

\begin{definition} \textbf{(constructions)}
\ \newline\vspace{-\baselineskip}
    \begin{enumerate}[label=(\roman*)]

    \item \textit{Variables x}, \textit{y}, ... are
        \textit{constructions} that construct objects (elements of
        their respective ranges) dependently on a valuation
        \textit{v}; they \textit{v-}construct.

    \item Where \textit{X} is an object whatsoever (even
        a construction), \textsuperscript{0}$\!$\textit{X} is the
        \textit{construction} \textit{Trivialization} that constructs
        \textit{X} without any change.

    \item Let \textit{X},\textit{
        Y}\textsubscript{1},...,\textit{Y\textsubscript{n}} be
        arbitrary constructions. Then \textit{Composition} [\textit{X
        Y}\textsubscript{1}...\textit{Y\textsubscript{n}}] is the
        following \textit{construction}. For any \textit{v}, the
        Composition [\textit{X
        Y}\textsubscript{1}...\textit{Y\textsubscript{n}}] is
        \textit{v-improper} if some of the constructions \textit{X},
        \textit{Y}\textsubscript{1},...,\textit{Y\textsubscript{n}} is
        \textit{v}{}-improper, or if \textit{X} does not
        \textit{v-}construct a function that is defined at the
        \textit{n-}tuple of objects \textit{v-}constructed by
        \textit{Y}\textsubscript{1},...,\textit{Y\textsubscript{n}}.
        If \textit{X} does \textit{v-}construct such a function then
        \hbox{[\textit{X
        Y}\textsubscript{1}...\textit{Y\textsubscript{n}}]}
        \textit{v-}constructs the value of this function at the
        \textit{n-}tuple.

    \item (${\lambda}${}-)\textit{Closure} [$\lambda
        $\textit{x}\textsubscript{1}...\textit{x\textsubscript{m}}\textit{
            Y}] is the following \textit{construction}. Let
            \textit{x}\textsubscript{1}, \textit{x}\textsubscript{2},
            ..., \textit{x\textsubscript{m}} be pair-wise distinct
            variables and \textit{Y} a construction. Then
            \hbox{[$\lambda $\textit{x}\textsubscript{1} ...
            \textit{x\textsubscript{m}}\textit{ Y}]}
            \textit{v}{}-\textit{constructs} the function \textit{f}
            that takes any members
            \textit{B}\textsubscript{1},...,\textit{B\textsubscript{m}}
            of the respective ranges of the variables
            \textit{x}\textsubscript{1},...,\textit{x\textsubscript{m}}
            into the object (if any) that is
            \textit{v}(\textit{B}\textsubscript{1}/\textit{x}\textsubscript{1},...,\textit{B\textsubscript{m}}/\textit{x\textsubscript{m}})-constructed
            by \textit{Y}, where
            \textit{v}(\textit{B}\textsubscript{1}/\textit{x}\textsubscript{1},...,\textit{B\textsubscript{m}}/\textit{x\textsubscript{m}})
            is like \textit{v} except for assigning
            \textit{B}\textsubscript{1} to
            \textit{x}\textsubscript{1}, ...,
            \textit{B\textsubscript{m}} to
            \textit{x\textsubscript{m}}.

    \item Where \textit{X} is an object whatsoever,
        \textsuperscript{1}$\!$\textit{X} is the \textit{construction}
        \textit{Execution} that \mbox{\textit{v-}constructs} what
        \textit{X v-}constructs. Thus if \textit{X} is
        a \textit{v-}improper construction or not a construction as
        all, \textsuperscript{1}$\!$\textit{X} is \textit{v-}improper.

    \item Where \textit{X} is an object whatsoever,
        \textsuperscript{2}$\!$\textit{X} is the \textit{construction
        Double} \textit{Execution}. If \textit{X} is not itself
        a construction, or if \textit{X} does not \textit{v-}construct
        a construction, or if \textit{X v-}constructs
        a \textit{v-}improper construction, then
        \textsuperscript{2}$\!$\textit{X} is \textit{v-}improper.
        Otherwise \textsuperscript{2}$\!$\textit{X v-}constructs what
        is \textit{v-}constructed by the construction
        \textit{v-}constructed by~\textit{X.} 

    \item Nothing is a \textit{construction}, unless it so follows
        from (i) through (vi).

    \end{enumerate}
\end{definition}
With constructions of constructions, constructions of functions,
functions, and functional values in our stratified ontology, we need
to keep track of the traffic between multiple logical strata. The
ramified type hierarchy does just that. The type of first-order
objects includes all objects that are not constructions. The type of
second-order objects includes constructions of first-order objects.
The type of third-order objects includes constructions of first- and
second-order objects. And so on, ad infinitum.
\begin{definition} \textbf{(ramified hierarchy of types)} Let \textit{B} be
a \textit{base}, where a base is a collection of pair-wise disjoint,
non-empty sets. Then:

\medskip
\noindent
    \textbf{T}\textbf{\textsubscript{1}}
    \textit{(types of order 1)}.
    \begin{enumerate}[label=\roman*)]

    \item Every member of \textit{B} is an elementary \textit{type of
        order 1 over B}.

    \item Let $\alpha $, $\beta $\textsubscript{1}, ..., $\beta
        $\textit{\textsubscript{m}} (\textit{m} {\textgreater} 0) be
        types of order 1 over \textit{B}. Then the collection ($\alpha
        $ $\beta $\textsubscript{1} ... $\beta
        $\textit{\textsubscript{m}}) of all \textit{m-}ary partial
        mappings from $\beta $\textsubscript{1} ${\times}$ ...
        ${\times}$ $\beta $\textit{\textsubscript{m}} into $\alpha
        $ is a functional \textit{type of} order 1\textit{ over B.}

    \item Nothing is a \textit{type of order 1 over B} unless it so
        follows from (i) and (ii).

    \end{enumerate}

\noindent
    \textbf{C}\textbf{\textit{\textsubscript{n}}}
    \textit{(constructions of order n)}

    \begin{enumerate}[label=\roman*)]

    \item Let \textit{x} be a variable ranging over a type of order
        \textit{n}. Then \textit{x} is a \textit{construction of order
        n over B.}

    \item Let \textit{X} be a member of a type of order \textit{n}.
        Then $^0\!X$, $^1\!X$, $^2\!X$ \ are \textit{constructions of
        order n over B.}

    \item Let \textit{X, X}\textsubscript{1},...,
        \textit{X\textsubscript{m}} (\textit{m} {\textgreater} 0) be
        constructions of order \textit{n} over \textit{B}. Then
        \hbox{[\textit{X X}\textsubscript{1}...\textit{X}\textit{\textsubscript{m}}]}
        is a \textit{construction of order n over B.}

    \item Let \textit{x}\textsubscript{1}, ...,
        \textit{x\textsubscript{m}}, \textit{X} (\textit{m}
        {\textgreater} 0) be constructions of order \textit{n} over
        \textit{B}. Then
        \hbox{[${\lambda}$\textit{x}\textsubscript{1}...\textit{x\textsubscript{m}}
        \textit{X}]} is a \textit{construction of order n over B}.

    \item Nothing is a \textit{construction of order n} \textit{over}
        \textit{B} unless it so follows from
        \textbf{C}\textbf{\textit{\textsubscript{n}}} (i)-(iv).

    \end{enumerate}

\pagebreak[3]
\noindent
    \textbf{T}\textbf{\textit{\textsubscript{n}}}\textbf{\textsubscript{+1}}
    (\textit{types of order n +} 1)

\noindent
    Let \tstar~be the collection of all
    constructions of order \textit{n} over \textit{B}. Then

    \begin{enumerate}[label=\roman*)]

    \item \tstar~and every type of order
        \textit{n} are \textit{types of order n +} 1.

    \item If \textit{m} {\textgreater} 0 and ${\alpha}$,
        ${\beta}$\textsubscript{1},...,${\beta}$\textit{\textsubscript{m}}
        are types of order \textit{n} + 1 over \textit{B}, then
        \hbox{(${\alpha}$ ${\beta}$\textsubscript{1} ...
        ${\beta}$\textit{\textsubscript{m}})} (see
        \textbf{T\textsubscript{1}} (ii)) is a \textit{type of order
        n} + 1 \textit{over B}.

    \item Nothing is a \textit{type of order n + 1 over B} unless it
        so follows from (i) and (ii).

    \end{enumerate}
\end{definition}
We model sets and relations by their characteristic functions. Thus,
for instance, ($o{\iota}$) is the type of a set of individuals, while
($o{\iota}{\iota}$) is the type of a relation-in-extension between
individuals. For the purposes of natural-language analysis, we are
assuming the following base of ground types:
\begin{quote}\hfuzz=11pt
    \begin{tabular}{ll}
        $o$: &the set of truth-values {\textbf{T, F}};\\

        $\iota $:  &the set of individuals (the universe of
        discourse);\\

        $\tau $:  &the set of real numbers (doubling as discrete
        times);\\

        $\omega $:  &the set of logically possible worlds (the logical
        space).\\
    \end{tabular}
\end{quote}
Empirical expressions denote \textit{empirical conditions} that may or
may not be satisfied at some world/time pair of evaluation. We model
these empirical conditions as possible-world-semantic (PWS)
\textit{intensions}.  PWS intensions are entities of type
(${\beta}{\omega}$): mappings from possible worlds to an arbitrary
type ${\beta}$. The type ${\beta}$ is frequently the type of
a \textit{chronology} of ${\alpha}${}-objects, i.e., a mapping of type
(${\alpha}{\tau}$). Thus ${\alpha}${}-intensions are frequently
functions of type ((${\alpha}{\tau}$)${\omega}$), abbreviated as
`${\alpha}$\textsubscript{${\tau}{\omega}$}'. \textit{Extensional
entities} are entities of a type ${\alpha}$ where \hbox{${\alpha}$
${\neq}$ (${\beta}{\omega}$)} for any type ${\beta}$. Where \textit{w}
ranges over ${\omega}$ and \textit{t} over ${\tau}$, the following
logical form essentially characterizes the logical syntax of empirical
language: \hbox{${\lambda}$\textit{w}${\lambda}$\textit{t}
[...\textit{w}...\textit{t}...]}. Examples of frequently used PWS
intensions are: \textit{propositions} of type
$o$\textsubscript{${\tau}{\omega}$}, \textit{properties of
individuals} of type ($o{\iota}$)\textsubscript{${\tau}{\omega}$},
binary \textit{relations-in-intension} between individuals of type
($o{\iota}{\iota}$)\textsubscript{${\tau}{\omega}$},
\textit{individual offices} (or \textit{roles}) of type
${\iota}$\textsubscript{${\tau}{\omega}$}, hyperintensional
\textit{attitudes} of type
($o{\iota}$\tstar)\textsubscript{${\tau}{\omega}$}.

Logical objects like \textit{truth-functions} and \textit{quantifiers}
are extensional: ${\wedge}$ (conjunction), ${\vee}$ (disjunction) and
${\supset}$ (implication) are of type ($ooo$), and ${\lnot}$
(negation) of type ($oo$). \textit{Quantifiers}
${\forall}$\textsuperscript{${\alpha}$},
${\exists}$\textsuperscript{${\alpha}$} are type-theoretically
polymorphous total functions of type ($o$($o{\alpha}$)), for an
arbitrary type ${\alpha}$, defined as follows. The \textit{universal
quantifier} ${\forall}$\textsuperscript{${\alpha}$} is a function that
associates a class \textit{A} of ${\alpha}${}-elements with \textbf{T}
if \textit{A} contains all elements of the type ${\alpha}$, otherwise
with \textbf{F}. The \textit{existential quantifier}
${\exists}$\textsuperscript{${\alpha}$} is a function that associates
a class \textit{A} of ${\alpha}${}-elements with \textbf{T} if
\textit{A} is a non-empty class, otherwise with \textbf{F}.

Below all type indications will be provided outside the formulae in
order not to clutter the notation. The outermost brackets of the
Closure will be omitted whenever no confusion can arise. Furthermore,
`\emph{X}/$\alpha$' means that an object \emph{X} is (a member) of
type $\alpha$. `X $\rightarrow _v \alpha$' means that \emph{X} is
typed to \emph{v}-construct an object of type $\alpha$, if any. We
write `X $\rightarrow \alpha$' if what is \emph{v}-constructed does
not depend on a valuation \emph{v}. Throughout, it holds that the
variables $w \rightarrow _v \omega$  and $t \rightarrow _v \tau$. If
$C\rightarrow _v \alpha _{\tau \omega}$  then the frequently
used Composition [[\emph{C} \emph{w}] \emph{t}], which is the
intensional descent (a.k.a. extensionalization) of the
$\alpha$-intension \emph{v}-constructed by \emph{C}, will be encoded
as `$C _{wt}$'. Whenever no confusion arises, we use traditional infix
notation without Trivialisation for truth-functions and the identity
relation, to make the terms denoting constructions easier to read.

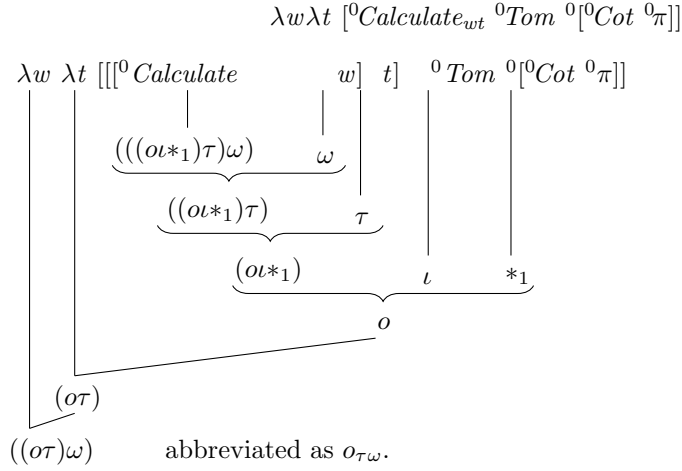
\begin{figure}
    \begin{center}
        ${\lambda}$\textit{w}${\lambda}$\textit{t}
        [$^0\!$\textit{Calculate\textsubscript{wt}}
        $^0\!$\textit{Tom}
        $^0$[$^0\!$\textit{Cot} $^0\! \pi$]]
    \end{center}

    \tikzset{ every node/.style={anchor=south west}, -}
    \begin{tikzpicture}
        \draw (0.5, 6.0) node {${\lambda}$\textit{w} ${\lambda}$\textit{t}   {[[[}\textsuperscript{0}\textit{Calculate}  \hspace{3em} \textit{w}{]}  ~\textit{t}]  \hspace{0.5em} \textsuperscript{0}\textit{Tom}   \textsuperscript{0}[\textsuperscript{0}$\!$\textit{Cot} \textsuperscript{0}${\pi}$]]};

        \draw (2.9, 6.1) -- (2.9,5.6);
        \draw (4.7, 6.1) -- (4.7,5.5);
        \draw [decorate,decoration={brace,mirror,amplitude=6pt}] (1.9,5.1) -- (5.0,5.1);
        \draw (1.8, 5.0) node {((($o{\iota}$\tstar[1])${\tau}$)${\omega}$)};
        \draw (4.5, 5.0) node {${\omega}$};

        \draw (5.2, 6.1) -- (5.2,4.7);
        \draw [decorate,decoration={brace,mirror,amplitude=6pt}] (2.5,4.3) -- (5.5,4.3);
        \draw (2.5, 4.2) node {(($o{\iota}$\tstar[1])${\tau}$)};
        \draw (5.0, 4.2) node {${\tau}$};

        \draw (6.1, 6.1) -- (6.1,3.9);
        \draw (7.2, 6.1) -- (7.2,3.9);
        \draw [decorate,decoration={brace,mirror,amplitude=6pt}] (3.5,3.5) -- (7.5,3.5);
        \draw (3.4, 3.4) node {($o{\iota}$\tstar[1])};
        \draw (5.9, 3.4) node {${\iota}$};
        \draw (7.0, 3.4) node {\tstar[1]};

        \draw (5.3, 2.8) node {$o$};

        \draw (1.4, 6.1) -- (1.4,2.3);
        \draw (5.4, 2.8) -- (1.4,2.3);
        \draw (1.0, 1.7) node {($o{\tau}$)};

        \draw (0.8, 6.1) -- (0.8,1.6);
        \draw (1.4, 1.8) -- (0.8,1.6);
        \draw (0.4, 1.0) node {(($o{\tau}$)${\omega}$) \hspace{2em} abbreviated as $o$\textsubscript{${\tau}{\omega}$}.};
    \end{tikzpicture}
    \caption{Analysis of the sentence ``\textit{Tom calculates
    cotangent of the number ${\pi}$}'' and its type derivation tree.}
    \label{fig:typeder}
\end{figure}

A simple example of the analysis of the sentence ``\textit{Tom
calculates cotangent of the number ${\pi}$}'' followed by its
derivation tree with type assignment is presented in
Figure~\ref{fig:typeder}.

The resulting type is the type of the proposition that Tom calculates
Cotangent of ${\pi}$. The types of the objects constructed by
\textsuperscript{0}$\!{\pi}$, $^0\!$\textit{Cot} and
[$^0\!$\textit{Cot} \textsuperscript{0}$\!{\pi}$], that is ${\tau}$,
(${\tau}{\tau}$) and ${\tau}$, respectively, are irrelevant here,
because these constructions are not constituents of the whole
construction. They occur only displayed by Trivialization
\textsuperscript{0}[$^0\!$\textit{Cot} \textsuperscript{0}$\!{\pi}$],
that is hyperintensionally. We are going to deal with this issue in
the next section.

\subsection{Context Recognition}

The algorithm of context recognition is based on definitional rules
presented in \cite{RefTIL2010}. Since these definitions are rather
complicated, here we introduce just the main principles which are
quite simple. TIL operates with a fundamental dichotomy between
hyperintensions (procedures) and their products, i.e. functions. This
dichotomy corresponds to two fundamental ways in which a construction
(meaning) can occur, to wit, \textit{displayed} or \textit{executed}.
If the construction is displayed then the \textit{construction} itself
becomes an object of predication (or an object to operate on); we say
that it occurs \textit{hyperintensionally}. If the construction occurs
in the execution mode, then it is a constituent of another
construction, and an additional distinction can be found at this
level. The constituent presenting a function may occur either
\textit{intensionally} or \textit{extensionally}. If intensionally,
then the whole \textit{function} is an object of predication (or to
operate on); if extensionally, then a functional \textit{value} is an
object of predication (to operate on). The two distinctions, between
displayed/executed and intensional/extensional occurrence, enable us
to distinguish between three kinds of context:
\begin{itemize}

\item \textit{hyperintensional context}: a construction occurs in
    a displayed mode though another construction at least one order
    higher needs to be executed in order to produce the displayed
    construction. In principle, constructions are displayed by
    Trivialization. It is important to realize that all the
    sub-constructions of a displayed construction occur also
    displayed.

\item \textit{intensional context}: a construction occurs in an
    executed mode in order to produce a \textit{function} but not its
    value; moreover, the executed construction does not occur within
    another hyperintensional context

\item \textit{extensional context}: a construction occurs in an
    executed mode in order to produce particular\textit{ value} of
    a function at a given argument; moreover, the executed
    construction does not occur within another intensional or
    hyperintensional context.

\end{itemize}
The basic idea underlying the above trifurcation is that the
\emph{same set of logical rules} applies to all three kinds of
context, but these rules operate on different complements:
\emph{procedures}, produced \emph{functions}, and functional
\emph{values}, respectively. A substitution is, of course, invalid if
something coarser-grained is substituted for something finer-grained.

The mechanism to display a construction is Trivialisation that raises
the context up to the hyperintensional level. However, we have to take
into account that Double Execution decreases the context down, because
$^{20}\! C$ is equivalent to \textit{C} in the sense of
\mbox{\textit{v-}constructing} the same object (or being
\textit{v-}improper) for the same valuations \textit{v.} According to
Def. 1 the Trivialisation $^0\! C$ constructs just the construction
\textit{C} which is afterwards executed. Thus $^{20}\! C$
\textit{v-}constructs the same object as does \textit{C}, or both
constructions are \textit{v-}improper.  Moreover, a higher-level
context is dominant over a lower-level one.  Thus, if \textit{C}
occurs in \textit{D} hyperintensionally, then all the subconstructions
of \textit{C} occur hyperintensionally in \textit{D} as well. If
\textit{C} occurs in the execution mode as a constituent of
\textit{D}, then the object that \textit{C v-}constructs (if any)
plays the role of an argument to operate on. In such a case, we have
to distinguish whether \textit{C} occurs \textit{intensionally} or
\textit{extensionally.} To this end we first distinguish between
extensional and intensional supposition of \textit{C}. Since
\textit{C} occurs executed, it is typed to\textit{ v}{}-constructs
a function \textit{f} of type
(${\alpha}{\beta}$\textsubscript{1}...${\beta}$\textit{\textsubscript{n}}),
\textit{n} possibly equal to zero. Now \textit{C} may be composed
within \textit{D} with constructions \textit{D}\textsubscript{1}, ...,
\textit{D\textsubscript{n}} which \textit{v-}construct the arguments
of the function \textit{f}, that is Composition [\textit{C}
\textit{D}\textsubscript{1}...\textit{D\textsubscript{n}}] is
a constituent of \textit{D.} In such a case we say that \textit{C}
occurs in \textit{D} with \textit{extensional supposition}. Otherwise
\textit{C} occurs in \textit{D} with intensional supposition that is
intensionally. Yet this is still not the whole story. If \textit{C}
occurs in \textit{D} with extensional supposition, it may still occur
intensionally, because Composition and ${\lambda}${}-Closure are dual
operations. While Composition decreases the context down, Closure
raises the context up to the intensional level. To take this issue
into account, we define a ${\lambda}${}-generic context induced by
a ${\lambda}${}-Closure. Then \textit{C} occurs extensionally within
\textit{D} if \textit{C} occurs with extensional supposition in
a non-generic context.

Figure~\ref{fig:conscontext} shows an example of the result of
automatic syntactic analysis of a construction including
context-recognition.

\begin{figure}
    \begin{center}
        \textbf{
        [[[}\textbf{\textsuperscript{0}$\!$}\textbf{\textit{Calculate w}}\textbf{]}\textbf{\textit{t}}\textbf{]} \textbf{\textsuperscript{0}}\textbf{\textit{Tom}}
        \textbf{\textsuperscript{0}}\textbf{[}\textbf{\textsuperscript{0}}$\!$\textbf{\textit{Cot}}
        \textbf{\textsuperscript{0}}\textbf{${\pi}$}\textbf{]]}

    \end{center}

    {\footnotesize
    \begin{tabbing}
    xxxx\=xx\=xx\=xx\=xx\=xx\=xx\=xx\=xx\=\kill
       \+{\textless}Composition context="INTENSIONAL"\\
        \>\>\>\>\>
        construction="\textbf{[[[}\textbf{\textsuperscript{0}}\textbf{\textit{Calculate
        w}}\textbf{]} \textbf{\textit{t}}\textbf{]}
        \textbf{\textsuperscript{0}}\textbf{\textit{Tom}}
        \textbf{\textsuperscript{0}}\textbf{[}\textbf{\textsuperscript{0}}\textbf{\textit{Cot}}
        \textbf{\textsuperscript{0}}\textbf{${\pi}$}\textbf{]]}"{\textgreater}\\

        \+{\textless}Composition
        context="EXTENSIONAL"
        construction="
        \textbf{[[}\textbf{\textsuperscript{0}}\textbf{\textit{Calculate
        w}}\textbf{]} \textbf{\textit{t}}\textbf{]}
        "{\textgreater} \\

        \+{\textless}Composition
        context="EXTENSIONAL"
        construction="\textbf{
        [}\textbf{\textsuperscript{0}}\textbf{\textit{Calculate w}}\textbf{]}
        "{\textgreater} \\

        {\textless}Trivialisation
        context="EXTENSIONAL"
        construction="
        \textbf{\textsuperscript{0}}\textbf{\textit{Calculate}}
        "/{\textgreater} \\

        \-{\textless}Variable
        context="INTENSIONAL"
        name="\textbf{\textit{w}}"/{\textgreater} \\

        {\textless}/Composition{\textgreater} \\

        \-{\textless}Variable
        context="INTENSIONAL"
        name="\textbf{\textit{t}}"/{\textgreater}\\

        {\textless}/Composition{\textgreater}\\

        {\textless}Trivialisation
        context="INTENSIONAL"
        construction="
        \textbf{\textsuperscript{0}}\textbf{\textit{Tom}}
        "/{\textgreater}
        \\

        \+{\textless}Trivialisation context="
        INTENSIONAL" construction="
        \textbf{\textsuperscript{0}}\textbf{[}\textbf{\textsuperscript{0}}\textbf{\textit{Cot}}
        \textbf{\textsuperscript{0}}\textbf{${\pi}$}\textbf{]}
        "{\textgreater} \\

        \+{\textless}Composition
        context="HYPERINTENSIONAL"
        construction="
        \textbf{[}\textbf{\textsuperscript{0}}\textbf{\textit{Cot}}
        \textbf{\textsuperscript{0}}\textbf{${\pi}$}\textbf{]}
        "{\textgreater} \\

        {\textless}Trivialisation
        context="HYPERINTENSIONAL"
        construction="
        \textbf{\textsuperscript{0}}\textbf{\textit{Cot}}
        "/{\textgreater} \\

        \-{\textless}Trivialisation
        context="HYPERINTENSIONAL"
        construction="\textbf{\textsuperscript{0}}\textbf{${\pi}$}"/{\textgreater} \\

        \-{\textless}/Composition{\textgreater} \\

        \-{\textless}/Trivialisation{\textgreater}\\

        {\textless}/Composition{\textgreater}\\

       \end{tabbing}
    }

    \caption{The result of automatic structural analysis of
    a construction with context-recognition.}
    \label{fig:conscontext}
\end{figure}

\subsection{Substitution Method and ${\beta}${}-Conversion by Value}

Having defined the three kinds of context, we are in a position to
specify an extensional logic of hyperintensions that flouts none of
the extensional logical principles like Leibniz's Law of substitution
of identicals or the rule of existential generalization. First, we
have defined valid rules of substitution. In an extensional context,
substitution is validated by so-called \textit{v}{}-congruent
constructions that \textit{v-}construct the same value of (possibly)
different functions.  In an intensional context, equivalent
constructions (constructions that are \textit{v}{}-congruent for every
valuation \textit{v} and thus \textit{v-}construct the same function)
are substitutable. In a hyperintensional context where the very
construction occurs as an argument, substitution is validated by
identical procedures that are procedurally isomorphic constructions.
But operating in a hyperintensional context \textit{requires} the
ability to operate directly on constructions, which is technically not
easy. To this end we have developed a well-tested method based on
substitution `\textit{by value}'. The technical complications we are
confronted with are rooted in \textit{displayed} constructions. For
instance, a variable occurring in a hyperintensional context is
displayed, i.e.  Trivialization-bound, which means being bound in
a manner that overrides ${\lambda}${}-binding. In particular, since
a displayed construction cannot at the same time be executed,
valuation does not play any role in such a context. Yet an argument of
the form
\begin{displaymath}
    \frac{\hbox{Tom calculates the cotangent of ${\pi}$}}
        {\hbox{Tom calculates the cotangent of something}}
\end{displaymath}
is obviously valid. In order to validly infer the conclusion, we need
to \textit{pre-process} the hyperintensional occurrence of the
Composition [\textsuperscript{0}$\!$\textit{Cot} \textit{x}] and
substitute the Trivialization of ${\pi}$ for \textit{x}. Only then can
the conclusion be inferred. In order to solve the problem, we deploy
the polymorphic functions
\textit{Sub\textsuperscript{n}}/(\tstar \tstar \tstar \tstar) and
\textit{Tr}\textsuperscript{${\alpha}$}/(\tstar
${\alpha}$) that operate on constructions in this manner.

The function \textit{Sub} when applied to constructions
\textit{C}\textsubscript{1}, \textit{C}\textsubscript{2} and
\textit{C}\textsubscript{3} returns as its value construction
\textit{D} that results from \textit{C}\textsubscript{3} by
substituting \textit{C}\textsubscript{1} for all occurrences of
\textit{C}\textsubscript{2} in \textit{C}\textsubscript{3}. The
function \textit{Tr} returns as its value the Trivialisation of its
argument.

For instance, let variable \textit{y} range over type ${\tau}$. Then
[\textsuperscript{0}\textit{Tr} \textit{y}]
\textit{v(}${\pi}$\textit{/y)}{}-constructs
\textsuperscript{0}${\pi}$. (Recall that
\textit{v(${\pi}$}\textit{/y)} is the valuation identical to
\textit{v} up to assigning the number ${\pi}$ to the variable
\textit{y.}) The Composition [\textsuperscript{0}$\!$\textit{Sub}
[\textsuperscript{0}$\!$\textit{Tr} \textit{y}]
\textsuperscript{0}$\!$\textit{x}
\textsuperscript{0}$\!$[\textsuperscript{0}$\!$\textit{Cot x}]]
\textit{v(}${\pi}$\textit{/y)}{}-constructs the Composition
[\textsuperscript{0}$\!$\textit{Cot} \textsuperscript{0}$\!$${\pi}$].

It should be clear now how to validly derive that Tom calculates the
cotangent of something if Tom calculates the cotangent of ${\pi}$. The
valid argument is this:
\begin{displaymath}
    \frac{\hbox{${\lambda}$\textit{w}${\lambda}$\textit{t}
    [[[\textsuperscript{0}\textit{Calculate w}]\textit{t}]
    \textsuperscript{0}\textit{Tom}
    \textsuperscript{0}[\textsuperscript{0}\textit{Cot}
    \textsuperscript{0}${\pi}$]]}}
    {\hbox{${\lambda}$\textit{w}${\lambda}$\textit{t}
    [\textsuperscript{0}${\exists}$ ${\lambda}$\textit{y}
    [[[\textsuperscript{0}\textit{Calculate w}]\textit{t}]
    \textsuperscript{0}\textit{Tom} [\textsuperscript{0}\textit{Sub}
    [\textsuperscript{0}\textit{Tr y}] \textsuperscript{0}\textit{x}
    \textsuperscript{0}[\textsuperscript{0}\textit{Cot}
    \textit{x}]]]]}}
\end{displaymath}
Existential quantifier ${\exists}$ is here a function of type
($o$($o{\tau}$)) that associates non-empty sets of numbers with the
truth-value T. The complete set of rules for quantifying into
hyperintensional attitudinal contexts has been specified and their
validity proved in \cite{HyperDJ}.

The substitution method is applied whenever we need to operate in
a hyperintensional context. But its application is much broader. In
particular, it is also applied in \textit{anaphora} pre-processing
(for details see \cite{DuziEJC2017}) and the specification of
${\beta}${}-conversion.

Though ${\beta}${}-conversion is the fundamental computational rule of
${\lambda}${}-calculi and functional programming languages, it is
underspecified by the commonly accepted rule
\begin{center}
    $[{\lambda}$\textit{x} \textit{C}(\textit{x}) \textit{A}$] \quad{}\vdash{}$
    \textit{C}(\textit{A}/\textit{x}).
\end{center}
The problem is this. Procedure of applying the function constructed by
\hbox{${\lambda}$\textit{x} \textit{C}(\textit{x})} to the argument
constructed by \textit{A} can be executed in two different, mutually
non-equivalent ways, to wit (a) by value or (b) by name.\footnote{See
also \cite{ChangFelleisen,IGPL,Plotkin}.} If by
name then the procedure \textit{A} is substituted for all the
occurrences of \textit{x} into \textit{C}. In this case there are two
problems. First, conversion of this kind is not guaranteed to be
a logically equivalent transformation as soon as partial functions are
involved. Second, it may yield loss of analytic information of which
function has been applied to which argument.\footnote{For the notion
of analytic information see \cite{DuziJPL}.} Strangely enough, purely
functional languages such as \textit{Clean} and \textit{Haskell} use
conversion by name. However, in our \textit{TIL-}based system
conversion \textit{by value} is applied. Its idea is simple. Execute
the procedure \textit{A} first, and only if \textit{A} does not fail
to produce an argument value on which \textit{C} is to operate,
substitute this value for \textit{x}. The rule of
${\beta}${}-conversion by value adapted to TIL is this.
\begin{center}
    [[${\lambda}$\textit{x}\textsubscript{1}...\textit{x\textsubscript{n}}
    \textit{Y}]
    \textit{D}\textsubscript{1}...\textit{D\textsubscript{n}}]
    ${}\rightarrow_\beta{}$
    \textsuperscript{2}[\textsuperscript{0}$\!$\textit{Sub}
    [\textsuperscript{0}$\!$\textit{Tr D}\textsubscript{1}]
    \textsuperscript{0}$\!$\textit{x}\textsubscript{1} ...
    [\textsuperscript{0}$\!$\textit{Sub} [\textsuperscript{0}$\!$\textit{Tr D}\textit{\textsubscript{n}}]
    \textsuperscript{0}$\!$\textit{x\textsubscript{n}}
    \textsuperscript{0}$\!$\textit{Y}]...]
\end{center}
This rule preserves logical equivalence, avoids the problem of loss of
analytic information, and moreover, in practice it is more efficient.
The efficiency is guaranteed by the fact that procedures
\textit{D}\textsubscript{1}, ..., \textit{D\textsubscript{n}} are
executed only once, whereas if these procedures are substituted for
all the occurrences of the \hbox{${\lambda}${}-bound} variables they
can subsequently be executed more than once.

For these reasons in our system we apply ${\beta}${}-conversion
\textit{by value.} The only exception is the so-called
\textit{restricted} ${\beta}${}-conversion by name that consists in
substituting variables for ${\lambda}${}-bound variables (ranging over
the same type). This is a technical simplification of a given
construction rather than the procedure of applying a function to its
argument. To adduce an example, we frequently apply the procedure of
\textit{extensionalization} of an intension. For instance, the
analysis of the sentence ``The Mayor of Ostrava is a computer
scientist'' comes down to two procedurally isomorphic constructions:
\begin{center}
    ${\lambda}$\textit{w}${\lambda}$\textit{t}
    [[\textsuperscript{0}$\!$\textit{Computer}
    \textsuperscript{0}$\!$\textit{Scientist}]\textit{\textsubscript{wt}}
    ${\lambda}w'\lambda t'$
    [\textsuperscript{0}$\!$\textit{Mayor\_of}$_{w't'}$
    \textsuperscript{0}$\!$\textit{Ostrava}]\textit{\textsubscript{wt}}]

\medskip

    ${\lambda}$\textit{w}${\lambda}$\textit{t}
    [[\textsuperscript{0}$\!$\textit{Computer}
    \textsuperscript{0}$\!$\textit{Scientist}]\textit{\textsubscript{wt}}
    [\textsuperscript{0}$\!$\textit{Mayor\_of\textsubscript{wt}}
    \textsuperscript{0}$\!$\textit{Ostrava}]]

\end{center}
The latter construction is a ${\beta}${}-reduced contractum of the
former.

\subsection{Partiality and presuppositions}

TIL is one of the few logics that deal with partial functions, see
also \cite{Moggi,Plotkin}. Partiality, as we all know well,
yields technical complications. But we have to work with partial
functions, because in an ordinary vernacular we use non-denoting yet
meaningful terms like `the King of France' or `the greatest prime'.
And reducing the domain of a partial function so that to obtain
a total function is not applicable here, because the domain reduction
often cannot be specified in a recursive way and we would end up with
a non-computable explosion of domains. There are two basic sources of
improperness.  Either a construction is a procedure of applying
a function \textit{f} to an argument \textit{a} such that \textit{f}
is not defined at \textit{a}, or a construction is type-theoretically
incoherent.  For instance, Composition
[\textsuperscript{0}$\!$\textit{Cotg} \textsuperscript{0}$\! \pi$] is
\textit{v-}improper for any valuation \textit{v}, because the function
cotangent is not defined at the number ${\pi}$ in the domain of real
numbers. \textit{Single Execution} \textsuperscript{1}$\!$\textit{X}
is improper for any valuation \textit{v} in case \textit{X} is not
a construction, because non-procedural object cannot be executed.

Sentences often come attached with a presupposition that is entailed
by the positive as well as negated form of a given sentence. Thus, if
a presupposition of a sentence \textit{S} is not true, the~sentence
\textit{S} can be neither true nor false. We follow Frege and Strawson
in treating survival~under negation as the most important test for
presupposition. Moreover, we take into account that there are two
kinds of~negation, namely Strawsonian \textit{narrow-scope} and
Russellian \textit{wide-scope negation}. While the~former is
presupposition-preserving, the latter is presupposition-denying.
Anyway, when dealing with sentences that come attached with
a presupposition, we need a general analytic schema for such sentences
that we are going to introduce now.

A sentence \textit{S} with a presupposition \textit{P} encodes as its
meaning this procedure:
\begin{center}
    In any ${\langle}$\textit{w, t}${\rangle}${}-pair of evaluation,

    \textit{if} \textit{P\textsubscript{wt}} is true

    \textit{then} evaluate \textit{S\textsubscript{wt}} to produce
    a truth-value,

    \textit{else fail} to produce a truth-value.

\end{center}
To formulate this schema rigorously, we need to define the
\textit{If-then-else-fail} function. Here is how.
\index{If-then-else!}The procedure encoded by ``If \textit{P}
(${\rightarrow}$ $o$) then \textit{C} (${\rightarrow}$ $\alpha $),
else \textit{D} (${\rightarrow}$ $\alpha $)'' behaves as follows:
\begin{enumerate}[label=\alph*)]

\item If \textit{P v-}constructs \textbf{T} then execute \textit{C}
    (and return the result of type ${\alpha}$, provided \textit{C} is
    not \textit{v-}improper)\textit{.}

\item If \textit{P v-}constructs \textbf{F} then execute \textit{D}
    (and return the result of type ${\alpha}$, provided \textit{D} is
    not \textit{v-}improper)\textit{.}
\item If \textit{P} is \textit{v-}improper then no result.

\end{enumerate}
Hence, \textit{If-then-else} is seen to be a function of type
$(\alpha o\tstar\tstar)$,
and its definition decomposes into two phases.

\textit{First}, select a construction to be executed on the basis of
a specific condition \textit{P}. The choice between \textit{C} and
\textit{D} comes down to this Composition:
\begin{center}
    [\textsuperscript{0}I\textsuperscript{*} ${\lambda}$\textit{c}
    [[\textit{P} ${\wedge}$ [\textit{c =}
    \textsuperscript{0}$\!$\textit{C}]] ${\vee}$ [${\lnot}$\textit{P}
    ${\wedge}$ [\textit{c =} \textsuperscript{0}$\!$\textit{D}]]]]
\end{center}
Types: \textit{P} ${\rightarrow}$\textit{\textsubscript{v}} $o$
\textit{v-}constructs the condition of the choice between the
execution of \textit{C} or \textit{D},
\textit{C}/\tstar,
\textit{D}/\tstar
${\rightarrow}$\textit{\textsubscript{v}} ${\alpha}$; \textit{c}
${\rightarrow}$\textit{\textsubscript{v}}
\tstar;
I\textsuperscript{*}/(\tstar($o$\tstar)):
the singularizer function (`the only one') that associates a singleton
of constructions with the construction that is the only element of
this singleton, and is otherwise (i.e. if the set is empty or
many-valued) undefined.

If \textit{P v-}constructs \textbf{T} then the variable \textit{c
v-}constructs the \textit{construction C}, and if \textit{P
v-}constructs \textbf{F} then the variable \textit{c v-}constructs the
\textit{construction D}. In either case, the set constructed by
\begin{center}
    ${\lambda}$\textit{c} [[\textit{P} ${\wedge}$ [\textit{c =}
    \textsuperscript{0}$\!$\textit{C}]] ${\vee}$ [${\lnot}$\textit{P}
    ${\wedge}$ [\textit{c =} \textsuperscript{0}$\!$\textit{D}]]]
\end{center}
is a singleton and the singularizer I\textsuperscript{*} returns as
its value either the construction \textit{C} or the construction
\textit{D.}\footnote{Note that in this phase \textit{C} and \textit{D}
are not constituents to be executed; rather they are merely displayed
as objects to be selected by the variable \textit{c.} This is to say
that in TIL constructions themselves can be objects to be operated on,
and without this \textit{hyperintensional} approach we would not be
able to define the \textit{strict} function \textit{if-then-else.}}

\pagebreak[2]
\textit{Second}, the selected construction is executed; therefore,
Double Execution must be applied:
\begin{center}
    \textbf{\textsuperscript{2}}[\textsuperscript{0}I\textsuperscript{*} ${\lambda}$\textit{c}
    [[\textit{P} ${\wedge}$ [\textit{c =} \textsuperscript{0}$\!$\textit{C}]]
    ${\vee}$ [${\lnot}$\textit{P} ${\wedge}$ [\textit{c =}
    \textsuperscript{0}$\!$\textit{D}]]]]
\end{center}
As a special case of \textit{P} being a presupposition, \textit{no}
construction \textit{D} is to be selected whenever \textit{P} is not
satisfied. Thus the definition of the \textit{if-then-else-fail}
function of type (${\alpha}o$\tstar) is this:
\begin{center}
    \textsuperscript{2}[\textsuperscript{0}I\textsuperscript{*}
    ${\lambda}$\textit{c} [\textit{P} ${\wedge}$ [\textit{c =}
    \textsuperscript{0}$\!$\textit{C}]]]
\end{center}
Now we can apply this definition to the case of a presupposition. Let
\hbox{\textit{P}/\tstar ${\rightarrow}$
$o$\textsubscript{${\tau}{\omega}$}} be a construction of
a presupposition of \textit{S}/\tstar
${\rightarrow}$ $o$\textsubscript{${\tau}{\omega}$}. Moreover, let
\hbox{\textit{c}/\tstar[n+1]
${\rightarrow}$\textit{\textsubscript{v}}
\tstar}, \mbox{\textsuperscript{2}\textit{c}
${\rightarrow}$\textit{\textsubscript{v}} $o$}. Then the type of the
\textit{if-then-else-fail} function is
($oo$\tstar) and its definition is:
\begin{center}
    ${\lambda}$\textit{w}${\lambda}$\textit{t}
    [\textsuperscript{0}\textit{if-then-else-fail
    P}\textit{\textsubscript{wt}}
    \textsuperscript{0}[\textit{S\textsubscript{wt}}]] =
    ${\lambda}$\textit{w}${\lambda}$\textit{t}
    \textsuperscript{2}[\textsuperscript{0}I\textsuperscript{*}
    ${\lambda}$\textit{c} [\textit{P\textsubscript{wt}}
    ${\wedge}$ [\textit{c =}
    \textsuperscript{0}[\textit{S\textsubscript{wt}}]]]]
\end{center}
\textit{Gloss.} In the first phase the construction
\textit{S\textsubscript{wt}} is selected, provided
\textit{P\textsubscript{wt}} \hbox{\textit{v-}constructs} \textbf{T}.
In the second phase \textit{S\textsubscript{wt}} is executed. In case
\textit{P\textsubscript{wt}} does not \hbox{\textit{v-}construct}
\textbf{T}, no construction is selected and executed, hence
\textsuperscript{2}[\textsuperscript{0}I\textsuperscript{*}
${\lambda}$\textit{c} [\textit{P\textsubscript{wt}} ${\wedge}$
\hbox{[\textit{c =}
\textsuperscript{0}[\textit{S\textsubscript{wt}}]]}]] is
\textit{v-}improper and the so constructed proposition has
a truth-value gap, as it should have.

In what follows, instead of the above definition we will use this
abbreviated notation as the \textit{general analytic schema}:
\begin{center}
    ${\lambda}$\textit{w}${\lambda}$\textit{t} [\textit{if}
    \textit{P\textsubscript{wt}} \textit{then}
    \textit{S\textsubscript{wt}} \textit{else} \textit{fail}].
\end{center}
For illustration, let us analyse Strawson's example \cite{Strawson1952}
\begin{center}
    All \textit{John's children} are asleep.
\end{center}
If the topic of the sentence is `John's children' then there is
a presupposition to the effect that John has children.\footnote{Hence
the situation is this. We are talking about John's children, and just
want to know what they are doing right now. The other option would be,
for instance, the scenario of talking about those who are asleep, and
the sentence would be offered as an answer, ``Among those who are
asleep are all of John's children''. On this reading the sentence
would merely entail but not presuppose that John has children.}  Hence
the truth-conditions of this reading can be formulated like this:
\begin{center}
    \textit{If} John has any children

    \textit{then} check whether each and every one of them is asleep

    \textit{else} fail to produce a truth-value.
\end{center}

\pagebreak[2]
\noindent
Thus, we have:
\begin{center}
    ${\lambda}$\textit{w}${\lambda}$\textit{t} [\textit{if}
    [\textsuperscript{0}${\exists}$
    [\textsuperscript{0}$\!$\textit{Children\_of\textsubscript{wt}}
    \textsuperscript{0}$\!$\textit{John}] \textit{then}
    [[\textsuperscript{0}$\!$\textit{All}
    [\textsuperscript{0}$\!$\textit{Children\_of\textsubscript{wt}}
    \textsuperscript{0}$\!$\textit{John}]]
    \textsuperscript{0}$\!$\textit{Sleep\textsubscript{wt}}]

    \textit{else} \textit{fail} ]
\end{center}
Types:
\textit{Children\_of}(($o{\iota}$)${\iota}$)\textsubscript{${\tau}{\omega}$}:
the empirical function (attribute) that dependently on a state of
affairs associates an individual with the set of those individuals who
are his or her children; \textit{John}/${\iota}$;
\textit{Sleep}/($o{\iota}$)\textsubscript{${\tau}{\omega}$};
${\exists}$/($o$($o{\iota}$));
\textit{All}/(($o$($o{\iota}$))($o{\iota}$)): restricted quantifier
that associates a set \textit{S} of individuals with all the superset
of \textit{S}.

\textit{Remark.} Here we use the restricted quantifier \textit{All},
because we want to arrive at the \textit{literal} analysis of the
sentence. Such an analysis follows Frege's principle
\cite{Frege1884}: It is simply not possible to
speak about an object without somehow denoting or naming
it.\footnote{The German original goes, ``Überhaupt ist es nicht
möglich von einem Gegenstand zu sprechen, ohne ihn irgendwie zu
bezeichnen oder benennen.''} If the unrestricted general quantifier
were used the resulting construction would be:
\begin{quote}
    ${\lambda}$\textit{w}${\lambda}$\textit{t} [if
    [\textsuperscript{0}${\exists}$
    [\textsuperscript{0}$\!$\textit{Children\_of\textsubscript{wt}}
    \textsuperscript{0}$\!$\textit{John}] then

    \hspace*{4em}[\textsuperscript{0}${\forall}{\lambda}$\textit{x}
    [[[\textsuperscript{0}$\!$\textit{Children\_of\textsubscript{wt}}
    \textsuperscript{0}$\!$\textit{John}] \textit{x}] ${\supset}$
    [\textsuperscript{0}$\!$\textit{Sleep\textsubscript{wt}}
    \textit{x}]]] else \textit{fail}]
\end{quote}
This is an equivalent construction producing the same proposition as
the above one, yet it is not the literal analysis of our sentence,
because the truth-function of implication is not mentioned in the
sentence.\footnote{For more details on the method of arriving at the
best literal meaning of a~sentence, see \cite{RefTIL2010}.}

\section{Automated Processing of Natural Language Input}
\label{sec:loganal}

Applying the Normal Translation Algorithm (NTA \cite{RefHorak2002}) to
natural language sentences allows the system to exploit the full
expressiveness of the natural language on the input side. In this
section, we first briefly describe the syntactic analysis part of the
NTA module, which builds the \emph{core} of the logical analysis, then
we show the main components of the logical construction building
process. Practically, the translation from natural language to logical
construction is implemented as a self-contained tool denoted as
AST\footnote{Automatic Semantic analysis Tool}
\cite{RefHorakMedved2015}, which can produce the logical analysis for
different input languages via specific language-dependent setup
files.\footnote{The current implementation can handle the input in the
Czech and English languages.}

\begin{figure}[t]
\begin{center}
    \forestset{node/.style={for tree={s sep=-6pt,l=7mm}}}
\begin{forest}node
[start
    [sentence
        [sentence
            [clause
                [np
                    [det
                        [\term{DET}{The}]
                    ]
                    [np
                        [\term{ADJ}{contractual}]
                        [np
                            [\term{N}{system}]
                        ]
                    ]
                ]
                [\term{V}{means}]
            ]
        ]
        [conj
            [\term{CONJ}{that}]
        ]
        [clause
            [np
                [det
                    [\term{DET}{the}]
                ]
                [np
                    [\term{N}{client}]
                ]
            ]
            [,phantom] [\term{V}{pays},l=9mm]
            [np
                [det
                    [\term{DET}{the}]
                ]
                [np
                    [\term{N}{company}]
                ]
            ]
            [np
                [det
                    [\term{DET}{a}]
                ]
                [np
                    [\term{ADJ}{monthly}]
                    [np
                        [\term{ADJ}{lump}]
                        [\term{N}{sum}]
                    ]
                ]
            ]
        ]
    ]
    [ends
        [\term{'.'}{.}]
    ]
]
\end{forest}
    \caption{Syntactic (phrasal) tree for ``\emph{The contractual
    system means that the client pays the company a  monthly lump
    sum.}''}
    \label{fig:tree}
\end{center}
\end{figure}
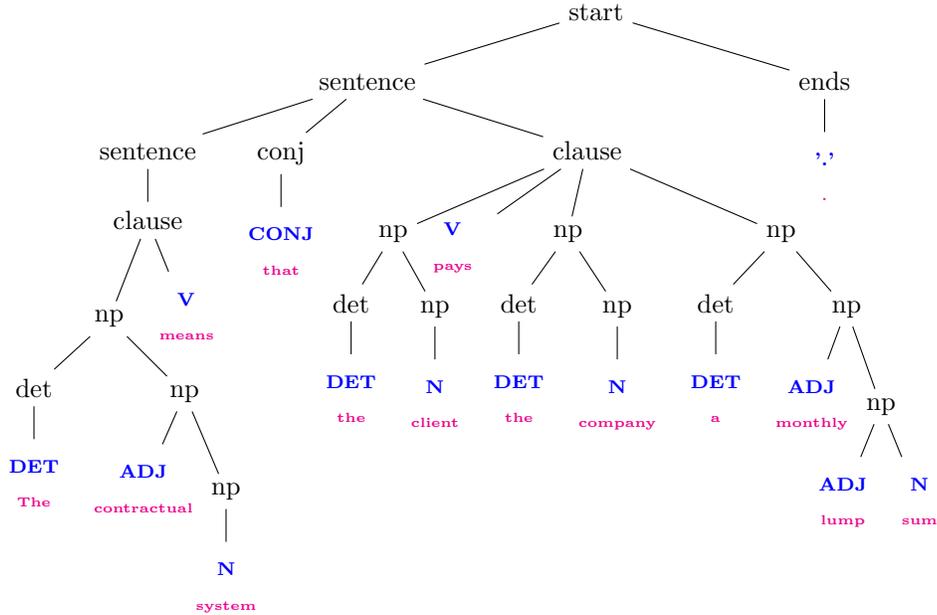

The semantic processing of a natural language sentence builds upon the
result of structural syntactic analysis or \emph{parsing}. As
a prevalent way of presenting the hierarchical organization of the
input sentence, most parsers are able to provide a comprehensive
representation in a form of a syntactic tree, which is also the form
processed by the AST tool.

The current version of the AST tool is able to process input in the
form of two basic types of syntactic trees -- a phrasal tree and
a dependency tree (see Figure~\ref{fig:tree} for an example).  The
employed parsers (the \texttt{synt}
parser \cite{RefHorak2008,RefJakubicek2009} and the SET
parser \cite{RefKovar2011}) are rule-based parsers with different
approaches to analysis: one is developed with the meta-grammar
concept, the core of the parser uses a context-free grammar with
contextual actions and it performs a stochastic agenda-based
head-driven chart analysis.  Its internal representation concentrates
on fast processing of very ambiguous syntactic structures. The parser
is able to process sentences with syntactic combinations resulting in
potentially thousands of possible syntactic trees ordered by a tree
score with an average processing time of 0.07\,s per sentence.

The second parser is also a rule-based one but it is based on
straightforward pattern-matching dependency rules. Its grammar
consists of a set of pattern matching specifications that compete with
each other in the process of dependency analysis. From the best
matches, the parser builds a full coverage syntactic dependency tree
of the input sentence. Currently, the system includes grammars for
Czech, Slovak and English, each with a few dozens of rules that
sufficiently model the syntax of the particular language, which is due
to the expressive character of the formalism.

The logical analysis then proceeds in interlinked modules of AST,
which process the output of a syntactic analyser and build the
resulting logical construction in a bottom-up manner. The core of AST
is language independent, but it needs four specific lexicons covering
the logical analysis of particular words and syntactic structures.

\subsection{The Syntactic-Semantic Grammar}

The bottom-up processing is driven by the hierarchical structure of
the input syntactic tree provided by the syntactic parser and by
a semantic extension of the actual grammar used in the parsing
process. To know which rule was used by the parser, AST needs
a semantic grammar specification, which contains the lists of semantic
actions that need to be done before propagation of particular node
constructions to the higher level in the syntactic tree. The semantic
actions define what logical functions correspond to each particular
syntactic rule. For instance, the \texttt{<np>} node (a \emph{noun
phrase}) in Figure~\ref{fig:tree} corresponds to the rule and action:
\begin{center}
\begin{minipage}{.5\textwidth}
    \begin{verbatim}
    np -> left_modif np
      rule_schema ( "[#1,#2]" )
    \end{verbatim}
\end{minipage}
\end{center}
which says that the resulting logical construction of the
\emph{left-hand side} \texttt{np} is obtained as a (logical)
application of the \texttt{left\_modif} (sub)construction to the
\emph{right-hand side} \texttt{np} (sub)construction. An example of
processing such grammar rule is in Figure~\ref{grammar_application}.

\begin{figure}[t]
\begin{alltt}
rule_schema: 2 nterms, '[#1,#2]'
1, 3, +np -> .{ left_modif } np . @level 0
  nterm 1: 1, 2, +left_modif -> .{ left_modifnl } . @level 0, k2eAgNnSc4
       TIL: \(\trZ{contractual}\)...\((\tw{(o\iota{})}\tw{(o\iota{})})\)
  nterm 2: 2, 3, +np -> .{ N } . @level 0, k1gNnSc4
       TIL: \(\trZ{system}\)...\(\tw{(o\iota{})}\)

Processing schema with params:
    #1: \(\trZ{contractual}\)...\((\tw{(o\iota{})}\tw{(o\iota{})})\)
    #2: \(\trZ{system}\)...\(\tw{(o\iota{})}\)
Resulting constructions:

    [\(\trZ{contractual}\)/\((\tw{(o\iota{})}\tw{(o\iota{})})\), \(\trZ{system}\)/\(\tw{(o\iota{})}\)]...\(\tw{(o\iota{})}\)
\end{alltt}
\caption{Analysis of the expression ``\textit{contractual system}''}
\label{grammar_application}
\end{figure}

\subsection{Typology of Lexical Items}

The second language dependent file defines lexical items and their TIL
types. The types are hierarchically built from four TIL ground
types \cite{RefTIL2010}.

AST contains rules for deriving implicit types based on Part-of-Speech
(PoS) categories of the input words, so as the lexicon must prescribe
the type only for cases that differ from the implicit definition. The
PoS categories are derived from the full morphological database
Majka \cite{smerk2007majka} and disambiguated with context-based
stochastic tagger Desamb \cite{desamb}. General approaches in this
respect can exploit unsupervised techniques such as \cite{singh2017},
however, the required exact logical object specification in AST makes
use of detailed PoS category information determined by means of the
large word forms database of the respective language.  A lexical item
example for the verb ``(to) pay'' is:
\begin{center}
    \begin{minipage}{.7\textwidth}
\begin{alltt}
pay
/k5/otriv \(((({o}({o}\tw{o})({o}\tw{o}))\omega)\iota\iota)\)
\end{alltt}
    \end{minipage}
\end{center}
The lexicon assigns the logical types based on PoS tag filters, i.e.
the type can differentiate for the exact category of the word within
the context of the input sentence.  In this example, the \textit{verb}
(category \texttt{k5}) is assigned a logical object via the
\emph{object-trivialisation} construction schema and the corresponding
logical type (here a \emph{verbal object with two $\iota$-arguments}
meaning ``to pay something to somebody'').

\subsubsection{Verb Subcategorization Features}

An important part of the predicate construction process consists in
determining the main verb and its arguments (subject, object, ...)
within the sentence. In some languages this process can be driven by
the word order, but an authoritative resource for this procedure
always needs to lean on detailed information about particular verb
subcategorization features of verb valencies.

The next language dependent lexicon is thus a file that defines verb
valencies and schema and type information for building the resulting
construction from the corresponding valency frame. An example for the
verb ``(to) pay'' is as follows
\begin{center}
    \begin{minipage}{.9\textwidth}\small
        \begin{alltt}
pay
hTc4-hPc3 :exists:V(v):V(v):and:V(v)=[[#0,try(#1),try(#2)],V(w)]
        \end{alltt}
    \end{minipage}
\end{center}
This record defines the valency of {\it $\langle$somebody$\rangle$
pays $\langle$something$\rangle$ to $\langle$somebody$\rangle$}, given
by the \emph{brief valency frame} expressions \texttt{hTc4} of the
\emph{object} (an inanimate noun phrase in accusative) and
\texttt{hPc3} of the \emph{patient} (an animate noun phrase in
dative), and the resulting construction of the \emph{verbal object}
(\texttt{V(v)}) derived as an application of the \emph{verb}
(\texttt{\#0}) to its arguments (the sentence objects) with possible
extensification (\texttt{try(\#1)} and \texttt{try(\#2)}) and the
appropriate possible world variable (\texttt{V(w)}).

\begin{figure}
    \begin{quote}
            $\lambda w_{1}\lambda t_{2}(\exists s_{1})\Biggl((\exists
    x_{3})(\exists
    i_{4})\Bigl(\bigl[\tr{Does}_{w_{1}t_{2}},i_{4},[\tr{Imp}_{w_{1}},x_{3}]\bigr]\
    {}\land{}x_{3}=[\tr{mean},s_{1}]_{w_{1}}{}\land{}$

    ${}\land{}\ \bigl[[\tr{contractual},\tr{system}]_{w_{1}t_{2}},i_{4}\bigr]\Bigr)\
    {}\land{}s_{1}=$

    $\trZcon{\Biggl[\lambda w_{3}\lambda t_{4}(\exists
    x_{5})(\exists i_{6})(\exists i_{7})(\exists
    i_{8})\biggl(\bigl[\tr{Does}_{w_{3}t_{4}},i_{8},[\tr{Imp}_{w_{3}},x_{5}]\bigr]\
    {}\land{}\ [\tr{company}_{w_{3}t_{4}},i_{6}]\ {}\land{}$

    ${}\land{}
    \Bigl[\bigl[\tr{monthly},[\tr{lump},\tr{sum}]\bigr]_{w_{3}t_{4}},i_{7}\Bigr]\
    {}\land{}\ x_{5}=[\tr{pay},i_{6},i_{7}]_{w_{3}}\ {}\land{}\
    [\tr{client}_{w_{3}t_{4}},i_{8}]\biggr)\Biggr]}\Biggr)\dots \tw{o}{}$

    \end{quote}
    \caption{An example of the resulting logical construction for the
input sentence ``\emph{The contractual system means that the client
pays the company a monthly lump sum.}''}
    \label{fig:logconstr}
\end{figure}

\subsubsection{Phrasal and Sentence Combinatory Expressions}

Last two lexicons involved in the AST logical analysis allow to
specify semantic schemata of combinations on the phrasal and sentence
level. For example, in the case of Czech the phrasal Combinatory
expressions include a list of semantic mappings of prepositional
phrases to valency expressions based on the head preposition.  The
file contains for each combination of a preposition and a grammatical
case of the included noun phrase all possible valency slots
corresponding to the prepositional phrase.

\pagebreak[3]

\noindent
For instance, the record for the preposition ``k'' (to) is displayed
as
\begin{center}
    \begin{minipage}{.3\textwidth}
        \begin{alltt}
k
3 hA hH
        \end{alltt}
    \end{minipage}
\end{center}
saying that ``k'' can introduce a datival prepositional phrase of
a \emph{where-to} direction \texttt{hA} (e.g.\ ``\emph{k~lesu}'' --
"to a forest"), or a modal \emph{how/what} specification \texttt{hH}
(e.g.\ ``\emph{k~večeři}'' -- ``to a dinner'').

The combinatory expressions on the sentence level are used when the
sentence structure contains subordination or coordination clauses.
The sentence schemata are classified by the conjunctions used between
clauses. An example for the conjunction ``but'' is:
\begin{center}
    \begin{minipage}{.7\textwidth}
    \begin{alltt}
    ("";"but") : "lwt(awt(#1) and awt(#2))"
    \end{alltt}
    \end{minipage}
\end{center}
The resulting construction builds a logical conjunction
(\emph{clause$_1$ \textbf{and} clause$_2$}) of the two clauses.

These lexicons and schema lists then drive the whole process of
standard translation of a natural language sentence into a structured
logical construction suitable for processing by the TIL inference
mechanism.

\section{Reasoning in TIL}
\label{sec:reason}

Having defined $\beta$-conversion by value, substitution method, the
\emph{If-then-else-fail} function and the rules for existential
quantification into hyperintensional contexts, we were in a position
to define the inference machine based on TIL. Tichý in
\cite{Tichy1982,Tichy1986} specified the deduction system for TIL by
applying sequent calculus.  Tichý's version was specified for pre-1988
TIL, i.e. TIL based on the simple theory of types. Duží extended this
version for TIL as of 2010, and formulated the results as an
extensional logic of hyperintensions in two papers \cite{Duz1,Duz2}.
In our system of questions, answers and reasoning over natural
language texts we decided to apply the general resolution method (GRM)
adjusted for TIL. There are two main reasons for this option. First,
deduction by applying the resolution method is goal/question driven.
And second, more importantly, this method is specified in an
algorithmic way and thus easy to implement. The first version was
implemented in Prolog \cite{EJC2009} and later extended to the
general resolution.

The idea to implement intelligent reasoning in Prolog or more
generally by the resolution method is not a new one. For instance,
Flach in~\cite{Flach1994} aims to make the reader familiar with
the implementation of the intelligent reasoning with natural language
using Prolog programs. Our novel contribution is both theoretical and
practical.  From the theoretical point of view, the novelty consists
in a fine-grained logical analysis of natural language in TIL, as
described above. In our opinion, in a multi-agent world of the
Semantic Web, information and communication technologies, artificial
intelligence, and other such facilities, there is a pressing need for
a universal framework informed by one logic making all the
semantically salient features of natural language explicit.  And TIL
with its procedural semantics and hyperintensional typing is such
a universal framework. From the practical point of view, we have been
developing an inference machine that neither over-infers (which yields
inconsistences) nor under-infers (which yields a lack of knowledge).

The transition from TIL into GRM and vice versa has been specified in
a near to equivalent way, i.e., without the loss of important
information encoded by TIL constructions. The algorithm of
transferring closed constructions that are typed to construct
propositions into the clausal form appropriate for GRM decomposes into
the following steps.

\begin{enumerate}[label=\roman*)]
    \item \textit{Eliminate the left-most}
        $\lambda$\emph{w}$\lambda$\emph{t}: \hspace{1pt} $[\lambda
        w \lambda t \, C(w,t)]  \Rightarrow C(w,t)$

    \item \textit{Eliminate unnecessary quantifiers that do not
        quantify any variable}

    \item \textit{Apply the rule of $\alpha$-conversion so that
        different $\lambda$-bound variables have different names}

    \item \textit{Eliminate truth functions} ${\supset}$ (implication)
        and ${\equiv}$ (equivalence) by applying these rules: [C
        ${\supset}$ D] ${}\vdash{}$ [${\lnot}$C ${\vee}$ D] and [C
        ${\equiv}$ D] ${}\vdash{}$ [[${\lnot}$C ${\vee}$ D] ${\wedge}$
        [${\lnot}$D ${\vee}$ C]]

    \item \emph{Apply de Morgan laws}

    \item If the construction contains existential quantifiers,
        eliminate them by applying \emph{Skolemization}

    \item Move \emph{general quantifiers to the left}

    \item \emph{Apply distributive laws}: \newline [[C $\wedge$ D]
        $\vee$ E] ${}\vdash{}$ [[C $\vee$ E] $\wedge$ [D $\vee$ E], [C
        $\vee$ [D $\wedge$ E]] ${}\vdash{}$ [[C $\vee$ D] $\wedge$ [C
        $\vee$ E]

    \end{enumerate}
The result is a construction in the clausal form proper for GRM.

\subsection{Example of Question Answering in TIL}

Here is an \emph{example} of the analysis of a few natural language
sentences and answering questions over this mini knowledge base.

\paragraph{Scenario.} Tom, Peter and John are members of a sport club.
Every member of the club is a skier or a climber. No climber likes
raining. All skiers like snow. Peter does not like what Tom likes, and
does like what Tom does not like. Tom likes snow and raining.

\smallskip
\noindent \emph{Question}: 	Is there in the club a sportsman who
is a climber but not a skier?  If so, who is it?

\paragraph{Analysis.}
\begin{flushleft}
\emph{Types}.
\textit{Tom, Peter, SC}/${\iota}$;
\textit{Member-of}/($o{\iota}{\iota}$)\textsubscript{${\tau}{\omega}$};
\textit{Skier}, \textit{Climber}/($o{\iota}$)\textsubscript{${\tau}{\omega}$}; \textit{Rain}, \textit{Snow}/${\alpha}$; \textit{Like}/($o{\iota}{\alpha}$)\textsubscript{${\tau}{\omega}$}.
\end{flushleft}
\begin{enumerate}[label=\alph*)]
\item  $\lambda w \lambda t $ $[[^0\!$\emph{Member-of}$\!_{wt}$ $^0\!$\emph{Tom} $^0\!$\emph{SC}$]$ $\wedge$
            $[^0\!$\emph{Member-of}$\!_{wt}$ $^0\!$\emph{Peter} $^0\!$\emph{SC}$]$ $\wedge$ \\
\hspace{5cm}$[^0\!$\emph{Member-of}$\!_{wt}$ $^0\!$\emph{John} $^0\!$\emph{SC}$]]$
\item  $\lambda w \lambda t $ $[^0{\forall} {\lambda} x \, [[^0\!$\emph{Member-of}$\!_{wt}$ $x$ $^0\!$\emph{SC}$]$ $\supset$ $[[^0\!$\emph{Skier}$_{wt}$ $x]$ $\vee$ $[^0\!$\emph{Climber}$_{wt} \, x]]]]$
\item  $\lambda w \lambda t \, [^0{\forall} {\lambda} x \, [[^0\!$\emph{Climber}$_{wt}$ $x]$ $ \supset $ ${\neg}[^0\!$\emph{Like}$_{wt}$ $x$ $^0\!$\emph{Rain}$]]]$
\item  $\lambda w \lambda t $ $[^0{\forall} {\lambda} x \, [[^0\!$\emph{Skier}$_{wt}$ $x$] $\supset$ $[^0\!$\emph{Like}$_{wt}$  $x$ $^0\!$\emph{Snow}$]]]$
\item  $\lambda w \lambda t $ $[^0 {\forall} {\lambda} x \, [[[^0\!$\emph{Like}$_{wt}$ $^0\!$\emph{Tom} $x]$ $\supset \neg[^0\!$\emph{Like}$_{wt}$ $^0\!$\emph{Peter} $x ]]$ $\wedge$ \\
\hspace{2cm} $[\neg [^0\!$\emph{Like}$_{wt}$ $^0\!$\emph{Tom} $x ]$ $\supset$ $[^0\!$\emph{Like}$_{wt}$ $^0\!$\emph{Peter} $x]]]]$
\item  $\lambda w \lambda t $ $[[^0\!$\emph{Like}$_{wt}$ $^0\!$\emph{Tom} $^0\!$\emph{Snow}$]$ $\wedge$ $[^0\!$\emph{Like}$_{wt}$ $^0\!$\emph{Tom} $^0\!$\emph{Rain}$]]$
\item[Q)]  $\lambda w \lambda t $ $[^0{\exists} {\lambda} x \, [[^0\!$\emph{Member-of}$\!_{wt}$ $x$ $^0\!$\emph{SC}$]$ $\wedge$ $[^0\!$\emph{Climber}$_{wt}$ $x ]$ $\wedge$ $\neg[^0\!$\emph{Skier}$_{wt}$ $x ]]]$
\end{enumerate}

\paragraph{Solution.}
\begin{enumerate}[label=\arabic*)]
\item eliminate left-most $\lambda w \lambda t$ + $\alpha$-conversion:

\begin{enumerate}[label=\alph*)]
\item $[[^0\!$\emph{Member-of}$\!_{wt}$ $^0\!$\emph{Tom} $^0\!$\emph{SC}$]$ $\wedge$
            $[^0\!$\emph{Member-of}$\!_{wt}$ $^0\!$\emph{Peter} $^0\!$\emph{SC}$]$ $\wedge$ \\
           $[^0\!$\emph{Member-of}$\!_{wt}$ $^0\!$\emph{John} $^0\!$\emph{SC}$]]$
\item $[^0{\forall} {\lambda} x \, [[^0\!$\emph{Member-of}$\!_{wt}$ $x$ $^0\!$\emph{SC}$]$ $\supset$ $[[^0\!$\emph{Skier}$_{wt}$ $x]$ $\vee$ $[^0\!$\emph{Climber}$_{wt} \, x]]]]$
\item
$[^0{\forall} {\lambda} y \, [[^0\!$\emph{Climber}$_{wt}$ $y]$ $\supset$ ${\neg}[^0\!$\emph{Like}$_{wt}$ $y$ $^0\!$\emph{Rain}$]]]$
\item  $[^0 {\forall} {\lambda} z \, [[^0\!$\emph{Skier}$_{wt}$ $z]$ $\supset$ $[^0\!$\emph{Like}$_{wt}$ $z$ $^0\!$\emph{Snow} $]]]$
\item
$[^0 {\forall} {\lambda} u \, [[[^0\!$\emph{Like}$_{wt}$ $^0\!$\emph{Tom} $u]$ $\supset \neg[^0\!$\emph{Like}$_{wt}$ $^0\!$\emph{Peter} $u ]]$ $\wedge$ \\
 $[\neg [^0\!$\emph{Like}$_{wt}$ $^0\!$\emph{Tom} $u ]$ $\supset$ $[^0\!$\emph{Like}$_{wt}$ $^0\!$\emph{Peter} $u]]]]$
\item  $[[^0\!$\emph{Like}$_{wt}$ $^0\!$\emph{Tom} $^0\!$\emph{Snow}$]$ $\wedge$ $[^0\!$\emph{Like}$_{wt}$ $^0\!$\emph{Tom} $^0\!$\emph{Rain}$]]$
\item[Q)]  $[^0{\exists} {\lambda} q \, [[^0\!$\emph{Member-of}$\!_{wt}$ $q$ $^0\!$\emph{SC}$]$ $\wedge$ $[^0\!$\emph{Climber}$_{wt}$ $q]$ $\wedge$ $\neg[^0\!$\emph{Skier}$_{wt} \, q]]]$
\end{enumerate}

\item negate the question Q)
\begin{enumerate}[label=\alph*)]
\item[G)]
$[^0{\forall} {\lambda} q \, [{\neg}[^0\!$\emph{Member-of}$\!_{wt}$ $q$ $^0\!$\emph{SC}$]$ $\vee$ $\neg[^0\!$\emph{Climber}$_{wt}$ $q$ $]$ $\vee$ $[^0\!$\emph{Skier}$_{wt} \, q]]]$
\end{enumerate}

\item eliminate $\forall$

\begin{enumerate}[label=\alph*)]
\item $[[^0\!$\emph{Member-of}$\!_{wt}$ $^0\!$\emph{Tom} $^0\!$\emph{SC}$]$ $\wedge$
            $[^0\!$\emph{Member-of}$\!_{wt}$ $^0\!$\emph{Peter} $^0\!$\emph{SC}$]$ $\wedge$ \\
           $[^0\!$\emph{Member-of}$\!_{wt}$ $^0\!$\emph{John} $^0\!$\emph{SC}$]]$
\item $[[^0\!$\emph{Member-of}$\!_{wt}$ $x$ $^0\!$\emph{SC}$]$ $\supset$ $[[^0\!$\emph{Skier}$_{wt}$ $x]$ $\vee$ $[^0\!$\emph{Climber}$_{wt} \, x]]]$
\item
$[[^0\!$\emph{Climber}$_{wt}$ $y]$ $\supset$ ${\neg}[^0\!$\emph{Like}$_{wt}$ $y$ $^0\!$\emph{Rain}$]]$
\item  $[[^0\!$\emph{Skier}$_{wt}$ $z]$ $\supset$ $[^0\!$\emph{Like}$_{wt}$ $z$ $^0\!$\emph{Snow}$]]$
\item
$[[[^0\!$\emph{Like}$_{wt}$ $^0\!$\emph{Tom} $u]$ $\supset \neg[^0\!$\emph{Like}$_{wt}$ $^0\!$\emph{Peter} $u]$ $\wedge$ \\
 $[\neg [^0\!$\emph{Like}$_{wt}$ $^0\!$\emph{Tom} $u ]$ $\supset$ $[^0\!$\emph{Like}$_{wt}$ $^0\!$\emph{Peter} $u]]]$
\item  $[[^0\!$\emph{Like}$_{wt}$ $^0\!$\emph{Tom} $^0\!$\emph{Snow}$]$ $\wedge$ $[^0\!$\emph{Like}$_{wt}$ $^0\!$\emph{Tom} $^0\!$\emph{Rain}$]]$
\item[G)]
$[{\neg}[^0\!$\emph{Member-of}$\!_{wt}$ $q$ $^0\!$\emph{SC}$]$ $\vee$ $\neg[^0\!$\emph{Climber}$_{wt}$ $q$ $]$ $\vee$ $[^0\!$\emph{Skier}$_{wt} \, q ]]$
\end{enumerate}

\item eliminate $\supset$ from b), c), d), e)

\begin{enumerate}[label=\alph*)]
\item $[[^0\!$\emph{Member-of}$\!_{wt}$ $^0\!$\emph{Tom} $^0\!$\emph{SC}$]$ $\wedge$
            $[^0\!$\emph{Member-of}$\!_{wt}$ $^0\!$\emph{Peter} $^0\!$\emph{SC}$]$ $\wedge$ \\
           $[^0\!$\emph{Member-of}$\!_{wt}$ $^0\!$\emph{John} $^0\!$\emph{SC}$]]$
\item
$[\neg[^0\!$\emph{Member-of}$\!_{wt}$ $x$ $^0\!$\emph{SC}$]$ $\vee$ $[[^0\!$\emph{Skier}$_{wt}$ $x]$ $\vee$ $[^0\!$\emph{Climber}$_{wt} \, x]]]$
\item
$[\neg [^0\!$\emph{Climber}$_{wt}$ $y]$ $\vee$ ${\neg}[^0\!$\emph{Like}$_{wt}$ $y$ $^0\!$\emph{Rain}$]]$
\item
$[\neg [^0\!$\emph{Skier}$_{wt}$ $z]$ $\vee$ $[^0\!$\emph{Like}$_{wt}$ $z$ $^0\!$\emph{Snow}$]]$
\item
$[[\neg [^0\!$\emph{Like}$_{wt}$ $^0\!$\emph{Tom} $u]$ $\vee$ $\neg[^0\!$\emph{Like}$_{wt}$ $^0\!$\emph{Peter} $u]$ $\wedge$ \\
 $[[^0\!$\emph{Like}$_{wt}$ $^0\!$\emph{Tom} $u ]$ $\vee$ $[^0\!$\emph{Like}$_{wt}$ $^0\!$\emph{Peter} $u]]]$
\item  $[[^0\!$\emph{Like}$_{wt}$ $^0\!$\emph{Tom} $^0\!$\emph{Snow}$]$ $\wedge$ $[^0\!$\emph{Like}$_{wt}$ $^0\!$\emph{Tom} $^0\!$\emph{Rain}$]]$
\item[G)]
$[{\neg}[^0\!$\emph{Member-of}$\!_{wt}$ $q$ $^0\!$\emph{SC}$]$ $\vee$ $\neg[^0\!$\emph{Climber}$_{wt}$ $q$ $]$ $\vee$ $[^0\!$\emph{Skier}$_{wt} \, q ]]$
\end{enumerate}

\item eliminate $\wedge$ from a), e), f); and write down the clauses
\begin{enumerate}[label=\alph*)]
\item[A1)] $[^0\!$\emph{Member-of}$\!_{wt}$ $^0\!$\emph{Tom} $^0\!$\emph{SC}$]$
\item[A2)] $[^0\!$\emph{Member-of}$\!_{wt}$ $^0\!$\emph{Peter} $^0\!$\emph{SC}$]$
\item[A3)] $[^0\!$\emph{Member-of}$\!_{wt}$ $^0\!$\emph{John} $^0\!$\emph{SC}$]$
\item[B)] $[\neg [^0\!$\emph{Member-of}$\!_{wt}$ $x$ $^0\!$\emph{SC}$]$ $\vee$ $[^0\!$\emph{Skier}$_{wt}$ $x]$ $\vee$ $[^0\!$\emph{Climber}$_{wt}$ $x]]$
\item[C)] $[\neg[^0\!$\emph{Climber}$_{wt}$ $y]$ $\vee$ $\neg [^0\!$\emph{Like}$_{wt}$ $y$ $^0\!$\emph{Rain}$]]$
\item[D)] $[\neg[^0\!$\emph{Skier}$_{wt}$ $z ]$ $\vee$ $[^0\!$\emph{Like}$_{wt}$ $z$ $^0\!$\emph{Snow}$]]$
\item[E1)] $[\neg[^0\!$\emph{Like}$_{wt}$ $^0\!$\emph{Tom} $u]$ $\vee$ $\neg[^0\!$\emph{Like}$_{wt}$ $^0\!$\emph{Peter} $u]]$
\item[E2)] $[[^0\!$\emph{Like}$_{wt}$ $^0\!$\emph{Tom} $u]$ $\vee$ $[^0\!$\emph{Like}$_{wt}$ $^0\!$\emph{Peter} $u]]$
\item[F1)] $[^0\!$\emph{Like}$_{wt}$ $^0\!$\emph{Tom} $^0\!$\emph{Snow}$]$
\item[F2)] $[^0\!$\emph{Like}$_{wt}$ $^0\!$\emph{Tom} $^0\!$\emph{Rain}$]$
\item[G)] $[\neg [^0\!$\emph{Member-of}$\!_{wt}$ $q$ $^0\!$\emph{SC}$]$ $\vee$ $\neg [^0\!$\emph{Climber}$_{wt}$ $q]$ $\vee$ $[^0\!$\emph{Skier}$_{wt}$ $q]]$
\end{enumerate}

\item goal driven resolution
\begin{enumerate}[label=\alph*)]
\item[R1)]	$\neg[^0\!$\emph{Climber}$_{wt}$ $^0\!$\emph{Tom}$]$ $\vee$ $[^0\!$\emph{Skier}$_{wt}$ $^0\!$\emph{Tom}$]$ \hfill 			G + A1, $^0\!$\emph{Tom}/$q$
\item[R11)]	$\neg[^0\!$\emph{Member-of}$_{wt}$ $^0\!$\emph{Tom} $^0\!$\emph{SC}$]$ $\vee$ $[^0\!$\emph{Skier}$_{wt}$ $^0\!$\emph{Tom}$]$     \hfill 		R1 + B
\item[R12)]	$[^0\!$\emph{Skier}$_{wt}$ $^0\!$\emph{Tom}$]$ \hfill 						R11) + A1)
\item[R13)]	$[^0\!$\emph{Like}$_{wt}$ $^0\!$\emph{Tom} $^0\!$\emph{Snow}$]$ \hfill	R12 + D, $^0\!$\emph{Tom}/$z$
\item[R14)]	$\neg [^0\!$\emph{Like}$_{wt}$ $^0\!$\emph{Peter} $^0\!$\emph{Snow}$]$ \hfill	R13) + E1), $^0\!$\emph{Snow}/$u$
\item[R2)]	$\neg [^0\!$\emph{Climber}$_{wt}$ $^0\!$\emph{Peter}$]$ $\vee$ $[^0\!$\emph{Skier}$_{wt}$ $^0\!$\emph{Peter}$]$ \hfill			G + A2, $^0\!$\emph{Peter}/$q$
\item[R21)]	$\neg [^0\!$\emph{Member-of}$_{wt}$ $^0\!$\emph{Peter} $^0\!$\emph{SC}$]$ $\vee$ $[^0\!$\emph{Skier}$_{wt}$ $^0\!$\emph{Peter}$]$ \hfill 		R2 + B
\item[R22)]	$[^0\!$\emph{Skier}$_{wt}$ $^0\!$\emph{Peter}$]$ \hfill 			R21 + A2)
\item[R23)]	$[^0\!$\emph{Like}$_{wt}$ $^0\!$\emph{Peter} $^0\!$\emph{Snow}$]$ \hfill		R22 + D, $^0\!$\emph{Peter}/$z$
\end{enumerate}
\end{enumerate}
\begin{flushleft}
YES \hfill								R14 + R23
\end{flushleft}
\emph{Second Question}; Who?
\begin{flushleft}
$[\lambda q [[^0\!$\emph{Like}$_{wt}$ $q$ $^0\!$\emph{Snow}$]$
$\wedge$ $\neg [^0\!$\emph{Like}$_{wt}$ $q$ $^0\!$\emph{Snow}$]]$
$^0\!$\emph{Peter}$]$ \hfill	$\lambda$-abstraction of Yes, $q
= ^0\!$\emph{Peter}
\end{flushleft}

\paragraph{Answer.} Yes, there is a sportsman in the club who is
a climber but not a skier. He is Peter.

\section{Conclusions}
\label{sec:concl}

In this paper, we described the hyperintensional system of reasoning
over natural language texts. The system makes use of a close
cooperation between computational linguistics and logic. We
concentrated on two main issues, which to the best of our knowledge
are not satisfactorily dealt with in current reasoning systems. First,
a fine-grained linguistic and logical analysis of questions and
underlying texts is a necessary condition for a high-quality reasoning
and answering over the texts. To this end we have applied the Normal
Translation Algorithm (NTA), which is a method that integrates logical
analysis of sentences with the linguistic approach to semantics. The
result of NTA is a corpus of 6,272 TIL constructions analyzed from
newspaper text sentences that serve as an input for TIL inference
machine. We have applied the procedural approach of TIL together with
the algorithm of context recognition in order to implement TIL
extensional logic of hyperintensions so that to be able to derive
inferential knowledge from explicit knowledge encoded in a wide-range
of natural-language resources.

The work is still in progress and the direction for future research is
clear. We plan to extend the coverage of the obtained techniques for
two languages, Czech and English so that to obtain a bi-lingual
system. Here we make use of the definition of procedural isomorphism.
Since we explicate structured meanings procedurally, any two terms or
expressions, even in different languages, are synonymous whenever they
are furnished with procedurally isomorphic constructions as meanings. Yet
the clarity of this direction does not imply its triviality. The
complexity of the work going into building such a system is almost
certain to guarantee that complications we are currently unaware of
will crop up. For instance, the rule of co-hyperintensionality, hence
of synonymy, has been formulated only conditionally. We define
a series of criteria for procedural isomorphism partially ordered
according to the degree of their being permissive with respect to
synonymy, from the strongest (most restrictive) to the weakest (most
liberal), depending on the area and language under scrutiny.
Furthermore, we provide good reasons for each of these criteria and
specify conditions under which this or that criterion is
applicable, for details, see \cite{Duzi-Synthese-2017}.

\section*{Acknowledgements}

This research has been supported by the Grant Agency of
the Czech Republic, project No. 18-23891S, ``Hyperintensional
Reasoning over Natural Language Texts'', and also by the
internal grant agency of VSB-TU Ostrava, project SGS No.
SP2018/172, ``Application of Formal Methods in Knowledge
Modelling and Software Engineering''.

\end{document}